\documentclass[letterpaper, 10 pt, conference]{ieeeconf}  
\usepackage{graphicx}
\usepackage{booktabs}
\usepackage{xcolor}
\usepackage{hyperref}
\usepackage{amsmath}
\usepackage{cleveref}
\usepackage{multirow}
\usepackage{placeins}
\usepackage{cuted}
\usepackage{caption}
\usepackage[all]{hypcap}
\usepackage{etoolbox}

\IEEEoverridecommandlockouts                    
\usepackage{amssymb}  

\newcommand{\ABBR}{\text{Agile-WAM}}

\title{\LARGE \bf
\ABBR{}: An Agile Tactile World Action Model for \\ Contact-Rich Robot Control
}

\author{Hanchu Zhou$^{1,\star}$, Brendan Lynch$^{2}$, Raman Goyal$^{2}$, Dechen Gao$^{1}$,\\ Begum Kasap$^{2}$, Boqi Zhao$^{1}$, Junshan Zhang$^{1}$
\thanks{$^{1}$Hanchu Zhou, Dechen Gao, Boqi Zhao, and Junshan Zhang are with University of California, Davis, One Shields Avenue, Davis, CA, USA.
        {\tt\small hczhou@ucdavis.edu} 
}%
\thanks{$^{2}$Raman Goyal, Begum Kasap, and Brendan Lynch are with Analog Devices, San Jose, CA, USA.
{\tt\small Brendan.Lynch@analog.com} 
(Corresponding author: Brendan Lynch)}%
\thanks{$^\star$Work done during an internship at Analog Devices.}
}

\begin{document}

\IEEEaftertitletext{\vspace{-2.5\baselineskip}}

\maketitle
\thispagestyle{empty}
\pagestyle{empty}

\begin{strip}
\centering
\includegraphics[width=0.95\textwidth]{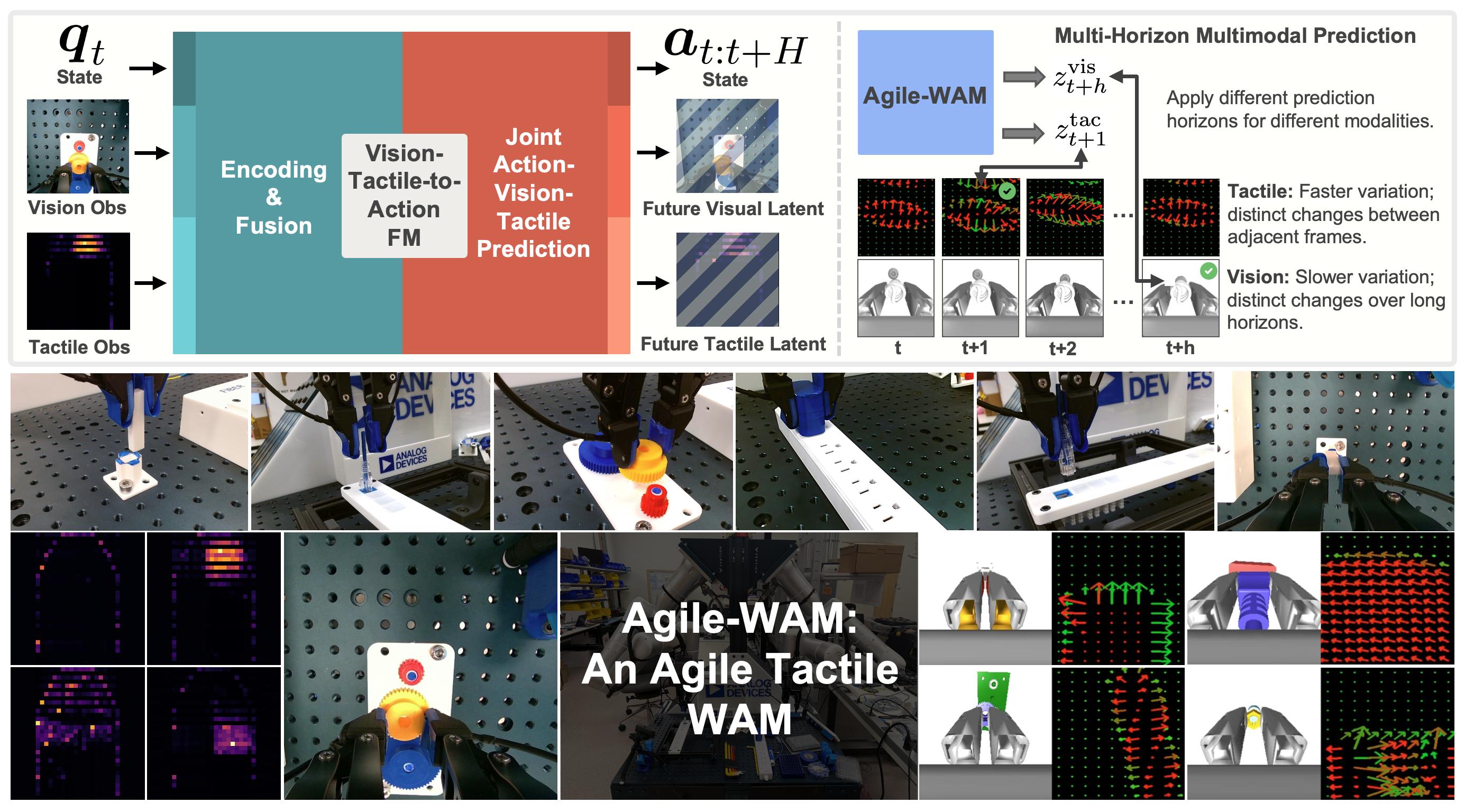}
\captionof{figure}{\ABBR{}, a tactile World Action Model,   takes multi-modal inputs  (including visual and tactile observations) and performs joint action--vision--tactile prediction through a direct vision-tactile-to-action flow-matching (FM) backbone.  To provide effective supervision for future latent prediction, \ABBR{} introduces multi-horizon multimodal prediction to enable joint prediction of visual and tactile latent over different horizons. By capturing the distinct temporal dynamics of the two modalities,  \ABBR{} aligns physical dynamics modeling better with action generation and enables more precise, fine-grained manipulation.
}
\label{fig:overview}
\vspace{0.15in}
\end{strip}

\begin{abstract}
World Action Models (WAMs) advance beyond conventional visuomotor policies by jointly predicting future world states and robot actions, enabling the policy to learn physical dynamics that support effective control. However,
 recent tactile WAMs often rely on large-scale pretrained generative backbones to capture contact-rich physical dynamics, which limit their inference efficiency and flexible deployment. In this paper, we present \ABBR{}, an agile tactile World Action Model for contact-rich robot control. \ABBR{} encodes visual and tactile observations into a shared latent  that serves as the source of a direct vision-tactile-to-action flow-matching process, which can jointly generate latent representations of  action chunks and future visual/tactile latents. A key observation is that vision and tactile signals evolve at inherently different timescales: adjacent visual frames are often highly similar, whereas tactile signals can change abruptly upon contact. We therefore introduce multi-horizon multimodal prediction in \ABBR{}, which provides supervision for visual latent at a larger temporal offset while predicting the tactile latent in the next frame to capture fine-grained contact dynamics. Across nine simulated and five real-world contact-rich manipulation tasks, \ABBR{} demonstrates strong and robust performance, outperforming the strongest baseline in success rate while maintaining low inference latency. In particular, in five real-world experiments, \ABBR{} yields a relative gain of $\textbf{29.4\%}$   in  overall success rates while achieving inference latency of $\textbf{11.9 ms}$. These results demonstrate that multimodal WAM can be achieved with an agile architecture suitable for precise and high-frequency robot control. More details are available on our \href{https://hanchuzhou.github.io/TARO_project_page/}{project page}.
\end{abstract}

\section{INTRODUCTION}


World Action Models (WAMs) have recently emerged as a promising paradigm for robot learning by coupling physical dynamics prediction with action generation. In contrast to conventional imitation learning or reinforcement learning paradigms \cite{gao2025ril}, WAMs  learn how the world will evolve and which actions should be executed jointly~\cite{guo2024prediction,LiS-RSS-25,wan2026worldagen,ye2026world}. By predicting future observations from consecutive frames, the WAM model is encouraged to capture dense spatiotemporal information about motion and interaction in the physical scene. Jointly modeling these future states, together with robot actions, further fosters the learned representation to connect physical dynamics evolution with action control. In this way, WAMs acquire a structured look-ahead capability and can produce actions that are more consistent with the anticipated evolution of the physical world.

Clearly, physical interaction is not fully observable from vision alone. This limitation is particularly pronounced in contact-rich manipulation, where visually similar observations may correspond to substantially different physical states, especially when the end effector contacts a rigid object with different levels of force. These tasks often require delicate and precise action adjustments despite exhibiting only subtle visual differences. Tactile sensing provides a complementary channel that captures local contact, force, and slip, and has consequently become increasingly important for learning precise manipulation policies~\cite{xue2025reactive,liu2025factr}.

Recent studies have begun to incorporate tactile sensing into world-action modeling by extending large pretrained video generation models with tactile observations and actions~\cite{yuan2026vtam,wu2026tactile, lou2026dream}. These early attempts demonstrate that explicitly modeling tactile dynamics can benefit contact-rich manipulation. However, reliance on large pretrained backbones introduces substantial computational and memory overhead during inference, which limits deployment on robotic platforms with constrained onboard resources and demanding high control frequencies. This gives rise to a fundamental question: \textit{Is it possible to develop an agile tactile-centric world-action model that enables flexible deployment across robotic platforms, while retaining the predictive benefits of world models?}

To tackle this challenge, we first note that visual and tactile observations  evolve at inherently different time scales, in the following sense: Visual appearance typically changes smoothly and slowly, making adjacent visual frames highly redundant, whereas tactile force field (TacFF) signals can change abruptly upon contact. It would be naive to apply an identical prediction horizon to both visual and tactile modalities, which would otherwise fail to account for their distinct temporal characteristics: Near-term visual observations often exhibit high redundancy, providing limited supervision for world modeling, while long-horizon tactile prediction may be less precise due to the rapid evolution of contact signals. We address this mis-alignment through \textbf{multi-horizon multimodal prediction}. Specifically, visual prediction is supervised at a larger temporal offset, where the observation exhibits more noticeable changes, encouraging the model to capture meaningful visual dynamics. In contrast, tactile prediction targets the next adjacent observation, allowing the model to track rapid contact changes without skipping fine-grained tactile transitions. This separate supervision mechanism better matches the temporal characteristics of each modality and supports precise contact-rich manipulation.

With this insight, we introduce \ABBR{}, an agile tactile World Action Model for contact-rich robot control. Rather than finetuning a large video foundation model, \ABBR{} employs a vision-tactile-to-action flow-matching backbone trained from scratch using visual observations and TacFF signals. The direct transition from vision-tactile to action obviates the need for repeated visual conditioning during the flow, thus significantly reducing the complexity \cite{gao2026vita}. As illustrated in \cref{fig:overview}, the model first encodes both visual and tactile modalities into fused latent representations, and then jointly predicts future visual observations and future tactile observations using multi-horizon multimodal prediction, as well as action chunks. Specifically, the fused latent representation, which captures both visual and tactile information, serves as the source distribution. A vision-tactile-to-action flow-matching network then learns a velocity field that directly transports this source representation toward a target latent composed of three parts: the action latent, future visual latent, and future tactile latent. We supervise these components with modality-specific objectives, applying the action loss in the action space and the visual and tactile prediction losses in their respective latent spaces. This joint objective encourages the policy representation to capture both visual evolution and local contact dynamics, thereby better aligning action generation with future dynamics and enabling fine-grained actions required for contact-rich manipulation. Meanwhile, the vision-tactile-to-action flow-matching backbone eliminates the need for costly conditioning mechanisms and enables a direct mapping from observations to actions and future observations using a compact backbone, thereby improving action precision and inference efficiency.

We evaluate \ABBR{} on nine simulated and five real-world contact-rich manipulation tasks, with focus on its performance and inference efficiency. Extensive ablations further isolate the contributions of tactile sensing and the multi-horizon multimodal prediction.

The main contributions of this paper are summarized as follows:
\begin{itemize}
    \item We introduce \ABBR{}, an agile tactile World Action Model that encodes visual and tactile multi-modal observations into a shared latent space, which directly drives a vision-tactile-to-action flow-matching model to jointly predict future visual observations, TacFF observations, and robot actions. In particular, the vision-tactile-to-action mechanism substantially reduces the computational overhead by removing cross-attention, thus improving inference efficiency. 

    \item We propose multi-horizon multimodal prediction that supervises visual and tactile signals at different temporal offsets, capitalizing  their inherent physical dynamics across distinct timescales, thus improving action generation for contact-rich manipulation.

    \item We evaluate \ABBR{} on nine simulated and five real-world contact-rich tasks. In particular,  in five real-world experiments,  \ABBR{} yields a relative gain of $29.4\%$ in overall success rates while achieving inference latency of $11.9\,\mathrm{ms}$. Comprehensive ablation studies demonstrate the effectiveness of each design choice and model component.

\end{itemize}

\section{Related Work}

\subsection{Tactile Integration in Robot Learning}

Tactile sensing provides complementary physical information on contact surface that is difficult to infer from vision alone. Recent work has developed transferable tactile representations across different sensors and modalities through self-supervised learning, multimodal alignment, and sensor-invariant representation learning~\cite{feng2025anytouch,gupta2025sensor,feng2026anytouch}. Beyond representation learning, tactile and force feedback have also been incorporated directly into manipulation policies to improve contact-rich control, with recent approaches emphasizing force-aware policy learning and fast tactile feedback for reactive manipulation~\cite{liu2025factr,xue2025reactive}. These studies demonstrate the importance of tactile feedback for physical interaction, but most use tactile signals primarily as additional policy inputs or reactive feedback. In contrast, \ABBR{} explicitly models the future evolution of tactile observations together with visual observations and actions, allowing tactile dynamics to directly participate in predictive policy learning.

\subsection{World Action Models for Robotics}

World Action Models extend conventional visuomotor policies by jointly learning future world evolution and robot actions, encouraging the policy to capture physical dynamics that are useful for control. Recent approaches have explored joint observation--action modeling~\cite{guo2024prediction,LiS-RSS-25,wan2026worldagen}. More recent work further scales this paradigm with large generative backbones, demonstrating that jointly modeling future observations and actions can improve physical generalization and support closed-loop robot control~\cite{ye2026world}. While most existing approaches focus primarily on visual observations, concurrent studies have begun incorporating tactile signals into predictive world-action modeling for contact-rich manipulation~\cite{yuan2026vtam,wu2026tactile}. However, these tactile extensions largely build upon large pretrained backbones. \ABBR{} instead studies an agile alternative that jointly models visual, tactile, and action dynamics with an efficient architecture designed for high-frequency robot control and low-resource deployment.

\subsection{Flow Matching for Generative Model and Robot Control}

Flow matching learns a continuous vector field that transports samples between source and target distributions, enabling high-quality generation with efficient inference and making it well suited for continuous action generation~\cite{lipman2022flow, zhang2024affordance}. In robot learning, flow-based policies have been applied to action generation from visual and 3D observations, while recent methods further improve inference efficiency by reducing the number of flow-matching steps required for action generation~\cite{chisari2024learning,hu2024adaflow,zhang2025flowpolicy}. Flow matching has also been successfully scaled to large generalist robot policies, demonstrating its effectiveness for learning complex and multimodal action distributions~\cite{black2024pi0}. More recent work develops noise-free flow-matching policies by replacing Gaussian noise with informative representations derived from observations or previous actions, achieving strong control performance with substantially improved inference efficiency~\cite{gao2026vita,zhang2025flowpolicy}. Inspired by this direction, \ABBR{} adopts vision-tactile-to-action direct flow matching as an efficient backbone and generalizes  it beyond action generation to jointly model future visual and tactile observations.

\section{METHODOLOGIES}

\begin{figure*}[t]
    \centering
    \includegraphics[width=1.0\linewidth]{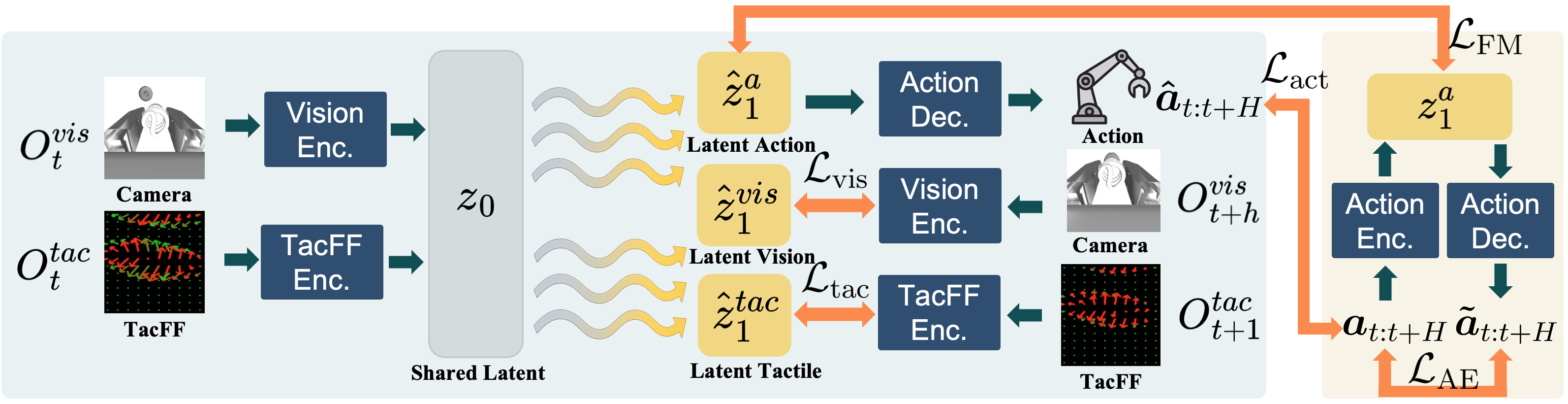}
    \caption{\textbf{Overview of \ABBR{}.} \ABBR{} encodes visual, tactile, and proprioceptive observations into a shared latent that serves as the source of a lightweight vision-tactile-to-action flow-matching backbone. The flow jointly generates latent representations of the action chunk and future visual and tactile observations. An action autoencoder provides a structured action latent space, while multi-horizon multimodal prediction uses a longer horizon for vision and a short horizon for tactile feedback to capture their distinct temporal dynamics. The model is trained end-to-end with flow-matching, action reconstruction and generation, and visual/tactile latent prediction losses.
}
    \label{fig:framework} 
    \vspace{-0.1in}
\end{figure*}

\ABBR{} aims to capture multimodal world dynamics in a shared latent space to support high-quality action generation. In this section, we describe how this objective is realized with a lightweight vision-tactile-to-action flow matching backbone and several key design choices. In~\cref{sec:joint_prediction}, we formulate the direct mapping from current observations to joint action-future predictions. In~\cref{sec:model_design}, we present the overall architecture of \ABBR{} and its training objectives. In~\cref{sec:multi_horzion_prediction}, we introduce multi-horizon multimodal prediction designed to better capture the distinct dynamics of different modalities.

\subsection{From Observation to Action--Vision--Tactile Joint Prediction}
\label{sec:joint_prediction}

We formulate robot control as joint action and future latent prediction. At environment step $t$, the robot receives a multimodal observation $O_t = \left[O_t^{\mathrm{vis}}, O_t^{\mathrm{tac}}, q_t\right]$, where $O_t^{\mathrm{vis}}$ denotes the visual observation, $O_t^{\mathrm{tac}}$ denotes the tactile force field (TacFF), and $q_t$ denotes robot’s proprioceptive states. Rather than learning only an action generation policy, our policy $\pi\left(a_{t:t+H}, z_t^{\mathrm{future}} \mid O_t\right)$ jointly models an action chunk and the future evolution of both modalities, where $a_{t+H}$ denotes an action chunk with prediction horizon $H$, of which only the first $h$ actions are executed before replanning. $z_t^{\mathrm{future}}=\left[z_{t+h_{\mathrm{vis}}}^{\mathrm{vis}}, z_{t+h_{\mathrm{tac}}}^{\mathrm{tac}} \right]$ comprises latent representations of future observations. We emphasize that jointly predicting future observations and actions encourages the learned representation to capture subtle physical dynamics that are directly relevant to contact-rich robot control, rather than relying only on the simple mapping from observation to action.

To this end, we develop this joint prediction algorithm using vision-tactile-to-action flow matching. Traditional flow-based policies typically start from a random Gaussian noise and repeatedly inject observations through conditioning modules, such as cross-attention. Instead, our backbone uses the latent representation of the current multimodal observation directly as the sole source of the flow, eliminating the need for an explicit conditioning mechanism during generation. This substantially simplifies the generation process and reduces computational overhead, leading to improved inference efficiency \cite{gao2026vita}.

We use learned encoders to encode the image and TacFF observations respectively, and fuse them into observation latent $z_0$ as the source of the flow. The target of the flow is divided into the action and future multimodal observations: $z_1 = \left[z_1^a,\, z_1^{\mathrm{vis}},\, z_1^{\mathrm{tac}} \right]$. The source and target are constructed to have the same dimensionality, as required by flow matching. For a flow time $\tau \in [0,1]$, we define the linear interpolation
\begin{equation*}
    z_\tau = (1-\tau)z_0 + \tau z_1,
\end{equation*}
whose target velocity is $z_1-z_0$. A lightweight flow network $v_\theta$ learns the vision-tactile-to-action velocity field through
\begin{equation*}
    \mathcal{L}_{\mathrm{FM}}
    =
    \mathbb{E}_{\tau,z_0,z_1}
    \left[
    \left\|
    v_\theta(z_\tau,\tau) - (z_1-z_0)
    \right\|_2^2
    \right].
\end{equation*}
Since sensory information is already embedded in the source $z_0$, $v_\theta$ does not require an additional observation-conditioning module during ODE integration.

At inference time, the current observation is encoded once into $z_0$, after which we solve
\begin{equation*}
    \frac{d z_\tau}{d\tau}
    =
    v_\theta(z_\tau,\tau),
    \qquad
    z_{\tau=0}=z_0,
\end{equation*}
from $\tau=0$ to $\tau=1$. This produces the predicted joint latent $\hat{z}_1=\left[\hat{z}_1^a,\, \hat{z}_1^{\mathrm{vis}},\, \hat{z}_1^{\mathrm{tac}}\right]$, from which the action chunk is decoded and executed.

\subsection{The Design of \ABBR{}}
\label{sec:model_design}

The overall architecture of \ABBR{} is illustrated in \cref{fig:framework}. The model consists of modality-specific encoders, an action autoencoder, and a lightweight flow-matching network.

\paragraph{Multimodal observation encoding}
At the input side, the learned visual encoder and tactile encoder extract representations from the current RGB image $O_t^{\mathrm{vis}}$ and TacFF $O_t^{\mathrm{tac}}$ signal respectively. Their features are concatenated with proprioceptive states $q_t$ and subsequently fused into the source latent $z_0$ through linear projection. In this way, visual and tactile information is incorporated once at the beginning of the flow, allowing the subsequent generation process to operate directly in the fused latent space without repeatedly injecting conditions.

\paragraph{Latent action representation}
Inspired by VITA ~\cite{gao2026vita}, we adopt the same design on action decoding that provides supervision on action space to prevent the collapse of action generation. We therefore introduce an action encoder $E_a$ and decoder $D_a$ to construct a structured latent action space:
\begin{equation*}
    z_1^a = E_a(a_{t:t+H}),
    \qquad
    \tilde{a}_{t:t+H} = D_a(z_1^a).
\end{equation*}
The action autoencoder is trained end-to-end with the policy using
\begin{equation}
    \mathcal{L}_{\mathrm{AE}}
    =
    \left\|
    a_{t:t+H}-D_a(E_a(a_{t:t+H}))
    \right\|_1.
    \label{eq:ae_loss}
\end{equation}

During inference, however, the action decoder receives the ODE-generated latent $\hat{z}_1^a$ rather than the encoder-generated target $z_1^a$. To align the decoder with the action latent generated via flow matching, we additionally decode the ODE-generated latent during training to provide supervision on action space:
\begin{equation}
    \mathcal{L}_{\mathrm{act}}
    =
    \left\|
    a_{t:t+H}-D_a(\hat{z}_1^a)
    \right\|_1.
    \label{eq:generated_action_loss}
\end{equation}

Gradients from this objective are propagated through the action decoder and the ODE integration process, directly anchoring the generated latent to executable ground-truth actions. Together, Eqs.~\eqref{eq:ae_loss} and~\eqref{eq:generated_action_loss} stabilize the learned action representation while reducing the discrepancy between training-time target latents and inference-time generated latents.

\subsection{Multi-horizon Multimodal Prediction}
\label{sec:multi_horzion_prediction}

A key challenge in multimodal world modeling is that different sensing modalities evolve at different temporal scales. Consecutive visual observations often exhibit substantial redundancy because scene appearance changes relatively smoothly. In contrast, tactile signals can vary sharply within a short period when the robot establishes contact or encounters resistance. Consequently, imposing the same prediction horizon on both modalities can lead to mismatched learning signals: a short visual horizon provides only trivial supervision due to the limited changes between nearby frames, whereas a long tactile horizon may blur the fine-grained contact dynamics essential for precise control.

To account for these heterogeneous dynamics, we introduce multi-horizon multimodal prediction. Instead of predicting vision and tactile latents at the same future step, their target latents are constructed as
\begin{equation*}
    z_1^{\mathrm{vis}}
    =
    E_{\mathrm{vis}}
    \left(O_{t+h_{\mathrm{vis}}}^{\mathrm{vis}}\right),
    \qquad
    z_1^{\mathrm{tac}}
    =
    E_{\mathrm{tac}}
    \left(O_{t+h_{\mathrm{tac}}}^{\mathrm{tac}}\right)
\end{equation*}
In our design, tactile prediction focuses on the next future frame, i.e., $h_{\mathrm{tac}}=1$, while visual prediction uses a longer temporal offset $h_{\mathrm{vis}}=h$ that matches the executed action length at inference. The longer visual horizon encourages the model to capture meaningful scene evolution rather than collapsing to duplicate current frames, whereas the shorter tactile horizon preserves abruptly changing contact information.

After solving the flow ODE, the predicted joint latent is decomposed into action, visual, and tactile components. Multimodal future prediction is supervised in latent space:
\begin{equation*}
    \mathcal{L}_{\mathrm{vis}}
    =
    \left\|
    \hat{z}_1^{\mathrm{vis}}
    -
    z_1^{\mathrm{vis}}
    \right\|_2^2,
    \quad
    \mathcal{L}_{\mathrm{tac}}
    =
    \left\|
    \hat{z}_1^{\mathrm{tac}}
    -
    z_1^{\mathrm{tac}}
    \right\|_2^2
\end{equation*}
Unlike pixel-level reconstruction, latent prediction provides a compact learning objective that encourages the shared flow representation to capture task-relevant evolution of both modalities without requiring expensive high-dimensional observation generation.

Together, the complete training objective is
\begin{equation*}
\begin{split}
    \mathcal{L}_{\ABBR}
    ={}&
    \lambda_{\mathrm{FM}}\mathcal{L}_{\mathrm{FM}}
    + \lambda_{\mathrm{AE}}\mathcal{L}_{\mathrm{AE}}
    + \lambda_{\mathrm{act}}\mathcal{L}_{\mathrm{act}}
    \\
    &+
    \lambda_{\mathrm{vis}}\mathcal{L}_{\mathrm{vis}}
    + \lambda_{\mathrm{tac}}\mathcal{L}_{\mathrm{tac}},
\end{split}
\end{equation*}
where $\lambda_{\mathrm{FM}}$, $\lambda_{\mathrm{AE}}$, $\lambda_{\mathrm{act}}$, $\lambda_{\mathrm{vis}}$, and $\lambda_{\mathrm{tac}}$ control the relative contributions of the five objectives. Through this joint optimization, \ABBR{} learns a compact flow representation that simultaneously captures multimodal world modeling and generates high-quality action sequences for contact-rich manipulation.

\section{EXPERIMENTS}

\begin{table*}[t]
\centering
\label{tab:sim_success_rates}
\caption{Success rates comparison on  simulation tasks}
\resizebox{\linewidth}{!}{
\begin{tabular}{lcccccc}
\toprule
\textbf{Task}
& \textbf{\ABBR{}}
& \textbf{VITA-VT}
& \textbf{VITA}
& \textbf{Tactile-WAM}
& \textbf{DP-VT}
& \textbf{DP} \\
\midrule
Bulb Screw             & \textbf{92.67}{\scriptsize$\pm$2.49} & 87.33{\scriptsize$\pm$0.94} & 71.33{\scriptsize$\pm$0.94} & 4.00{\scriptsize$\pm$0.00} & 4.67{\scriptsize$\pm$2.49} & 5.33{\scriptsize$\pm$0.94} \\
Gear Assembly          & \textbf{70.67}{\scriptsize$\pm$2.49} & 68.67{\scriptsize$\pm$1.89} & 68.33{\scriptsize$\pm$0.47} & 68.00{\scriptsize$\pm$1.63} & 58.67{\scriptsize$\pm$6.18} & 58.67{\scriptsize$\pm$1.89} \\
Power Plug Insertion   & 59.33{\scriptsize$\pm$0.94} & \textbf{63.33}{\scriptsize$\pm$3.27} & 54.67{\scriptsize$\pm$3.40} & 61.33{\scriptsize$\pm$2.49} & 58.00{\scriptsize$\pm$3.27} & 53.67{\scriptsize$\pm$1.25} \\
Peg Reorientation      & \textbf{42.00}{\scriptsize$\pm$1.63} & 28.67{\scriptsize$\pm$0.94} & 41.33{\scriptsize$\pm$4.11} & 34.67{\scriptsize$\pm$2.49} & 28.00{\scriptsize$\pm$2.83} & 22.00{\scriptsize$\pm$1.63} \\
Peg Insertion          & \textbf{49.33}{\scriptsize$\pm$0.94} & 47.33{\scriptsize$\pm$2.49} & 44.00{\scriptsize$\pm$1.67} & 40.00{\scriptsize$\pm$9.80} & 20.67{\scriptsize$\pm$13.20} & 21.33{\scriptsize$\pm$18.86} \\
USB Insertion          & \textbf{62.00}{\scriptsize$\pm$5.89} & 59.33{\scriptsize$\pm$9.43} & 55.33{\scriptsize$\pm$4.11} & 44.67{\scriptsize$\pm$8.06} & 38.00{\scriptsize$\pm$2.83} & 41.33{\scriptsize$\pm$5.25} \\
Ball Sorting           & \textbf{92.00}{\scriptsize$\pm$0.00} & 90.00{\scriptsize$\pm$3.27} & 72.00{\scriptsize$\pm$7.12} & 45.33{\scriptsize$\pm$0.94} & 74.67{\scriptsize$\pm$5.25} & 38.67{\scriptsize$\pm$3.40} \\
Object Search          & \textbf{50.67}{\scriptsize$\pm$2.49} & 36.00{\scriptsize$\pm$2.83} & 40.00{\scriptsize$\pm$3.27} & 33.33{\scriptsize$\pm$2.49} & 13.33{\scriptsize$\pm$0.94} & 14.67{\scriptsize$\pm$3.40} \\
Nut Bolt Threading     & \textbf{92.67}{\scriptsize$\pm$2.49} & 92.00{\scriptsize$\pm$2.83} & 86.67{\scriptsize$\pm$4.71} & 13.33{\scriptsize$\pm$2.49} & 72.00{\scriptsize$\pm$3.27} & 4.67{\scriptsize$\pm$0.94} \\
\bottomrule
\end{tabular}
}
\end{table*}

\begin{table}[t]
\centering
\caption{Success rate comparison on real-world tasks.}
\label{tab:real_success_rates}
\resizebox{\linewidth}{!}{
\begin{minipage}{\linewidth}
\centering

{\setlength{\tabcolsep}{4pt}
\begin{tabular}{lccc}
\toprule
\textbf{Policy}
& \textbf{Gear Assembly}
& \textbf{Peg Insertion}
& \textbf{Power Plug Insertion} \\
\midrule
\ABBR{}
& \textbf{0.80 (16/20)}
& 0.55 (11/20)
& \textbf{0.65 (13/20)} \\

VITA-VT
& 0.55 (11/20)
& \textbf{0.60 (12/20)}
& 0.35 (7/20) \\

VITA
& 0.75 (15/20)
& 0.25 (5/20)
& 0.25 (5/20) \\
\bottomrule
\end{tabular}
}

\vspace{0.5em}

{\setlength{\tabcolsep}{4.5pt}
\begin{tabular}{lcc}
\toprule
\textbf{Policy}
& \textbf{Internet Cable Insertion}
& \textbf{Internet Cable Unplugging} \\
\midrule
\ABBR{}
& \textbf{0.30 (6/20)}
& \textbf{0.85 (17/20)} \\

VITA-VT
& 0.15 (3/20)
& 0.75 (15/20) \\

VITA
& 0.20 (4/20)
& 0.20 (4/20) \\
\bottomrule
\end{tabular}
}

\vspace{-1.0em}

\end{minipage}
}
\end{table}

We evaluate \ABBR{} on nine simulated and five real-world contact-rich manipulation tasks. The simulation experiments are conducted using ManiFeel \cite{luu2025manifeel}, a comprehensive benchmark for tactile manipulation policy learning that provides realistic tactile simulation and challenging tasks in which tactile feedback is essential. The platform uses a 7-DoF Franka Emika Panda robot equipped with a TacFF sensor on its gripper. The TacFF sensor has a resolution of $10 \times 14$, with each sensing point measuring force magnitude and directions along the $x$- and $y$-axes, resulting in a $10 \times 14 \times 3$ tactile observation. For each task, we use 20--50 demonstrations from the official dataset for training. Following the benchmark's standard setup, only the wrist camera of $256 \times 256$ is used during training and evaluation. Because the wrist camera view is frequently occluded as the gripper interacts with the target object, this setting highlights the importance of tactile feedback.

\begin{figure}[h]
    \centering
    \includegraphics[width=1.0\linewidth]{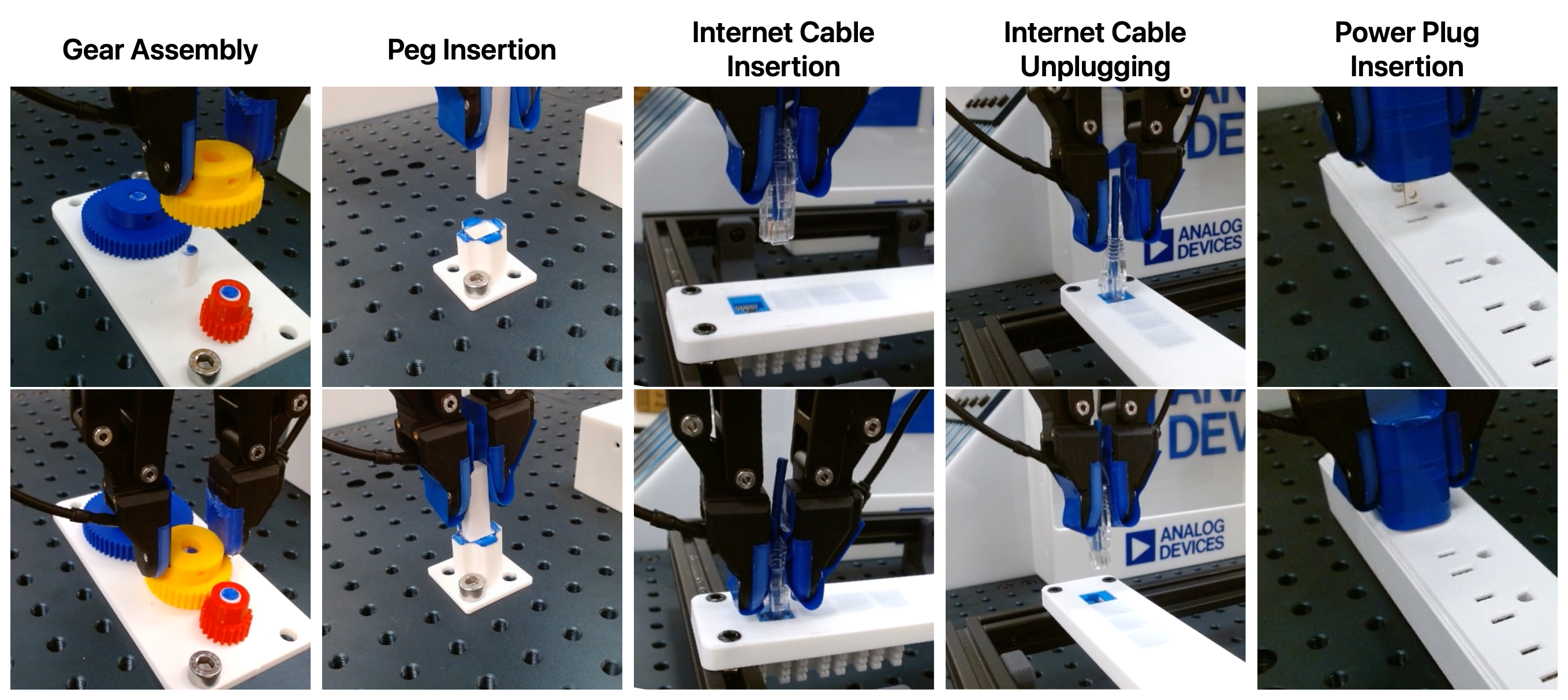}
    \caption{Real-world experiments include five challenging tasks: Gear Assembly, Peg Insertion, Internet Cable Insertion, Internet Cable Unplugging, and Power Plug Insertion.
    }
    \label{fig:task_overview} 
    \vspace{-0.1in}
\end{figure}

For the real-world experiments, we use a 7-DoF Flexiv Rizon 4 robot to evaluate five challenging tasks: Gear Assembly, Peg Insertion, Internet Cable Insertion, Internet Cable Unplugging, and Power Plug Insertion, as illustrated in \cref{fig:task_overview}. Visual observations are captured using an Intel RealSense D405 RGB-D wrist camera at a resolution of $320 \times 240$ and a frame rate of 30 FPS. Tactile feedback is acquired at 30 Hz using an Analog Devices $32 \times 32$ piezoresistive pressure sensor mounted on the gripper. For each task, we collect 50 expert demonstrations through teleoperation using a Meta Quest 3 headset.

\subsection{Experiment Settings}

\textbf{Backbones.}
We use an ImageNet-pretrained ResNet-18 as the visual encoder and an MLP-based autoencoder to encode and reconstruct action sequences. Because the TacFF signal has a spatial structure analogous to that of an image, we employ a separate ResNet-18, trained from scratch, as the tactile encoder to extract contact-related features. For flow matching, we adopt a lightweight MLP to parameterize the velocity field. During inference, we integrate the learned ODE using an explicit Euler solver with 6 steps to generate action and future latents.

\textbf{Baselines.}
We compare \ABBR{} with state-of-the-art vision-based and tactile-aware policies, including Diffusion Policy with visual observations (DP), Diffusion Policy with visual and tactile observations (DP-VT) \cite{luu2025manifeel}, Tactile-WAM \cite{wu2026tactile}, and the  vision-only (VITA) \cite{gao2026vita} and vision--tactile variants of VITA (VITA-VT). For a fair comparison, we reproduce Tactile-WAM by incorporating its core component, the Tactile Asymmetric Attention mechanism, into the tactile diffusion-policy implementation provided by ManiFeel. We additionally construct the VITA-VT by augmenting the original vision-only VITA with tactile observations using the same multimodal fusion strategy as \ABBR{}. The performance and efficiency comparisons are presented in \cref{sec:performance}.

\textbf{Real-World Tasks.}
We design five challenging real-world manipulation tasks that require tactile feedback for precise object alignment and correction. As illustrated in \cref{fig:task_overview}, the tasks are:

\textit{Gear Assembly:} The gear set is 3D-printed using the same assets as the corresponding ManiFeel task. The robot must insert the middle gear onto its shaft and then perform slight rotation until its teeth properly engage with the neighboring gears.

\textit{Peg Insertion:} The peg and base are 3D-printed from the ManiFeel assets. The robot must align the peg with the hole and insert it successfully.

\textit{Internet Cable Insertion:} The robot inserts a standard Ethernet cable with a locking tab into the port. It must carefully align the port and push it in until it locks.

\textit{Internet Cable Unplugging:} The robot must grasp the Ethernet cable, depress the locking tab to release the connector, and then pull the cable out of the port.

\textit{Power Plug Insertion:} The robot must align and insert a two-prong power plug into a surge protector. As the plug approaches the outlet, it occludes most of the wrist-camera view, making tactile feedback particularly important.

\textbf{Training and Evaluation.}
In both simulation and real-world settings, all methods use an action chunk horizon $H=16$ and execute $h=8$ actions at each inference step. We use 6 ODE integration steps for VITA, VITA-VT, and \ABBR{}, and 100 denoising steps for DP and DP-VT. During training, we evaluate each policy every 500 training steps with 50 evaluation rollouts in simulation. Policies are trained for 40k--100k steps to ensure convergence of the success rate. For each task, we report the highest success rate achieved during training, averaged over three random seeds. In the real-world experiments, each policy is evaluated for 20 episodes per task. All training and evaluation experiments can be performed on a single NVIDIA RTX 4090 GPU.

\subsection{Performance}
\label{sec:performance}

In this section, we highlight the performance of \ABBR{} that it matches or outperforms the state-of-the-art methods with respect to success rate. Additionally, we demonstrates its competitive inference efficiency against baseline while retain the benefit of world modeling.

\subsubsection{Success Rates}
\begin{figure}[h]
    \centering
    \includegraphics[width=1.0\linewidth]{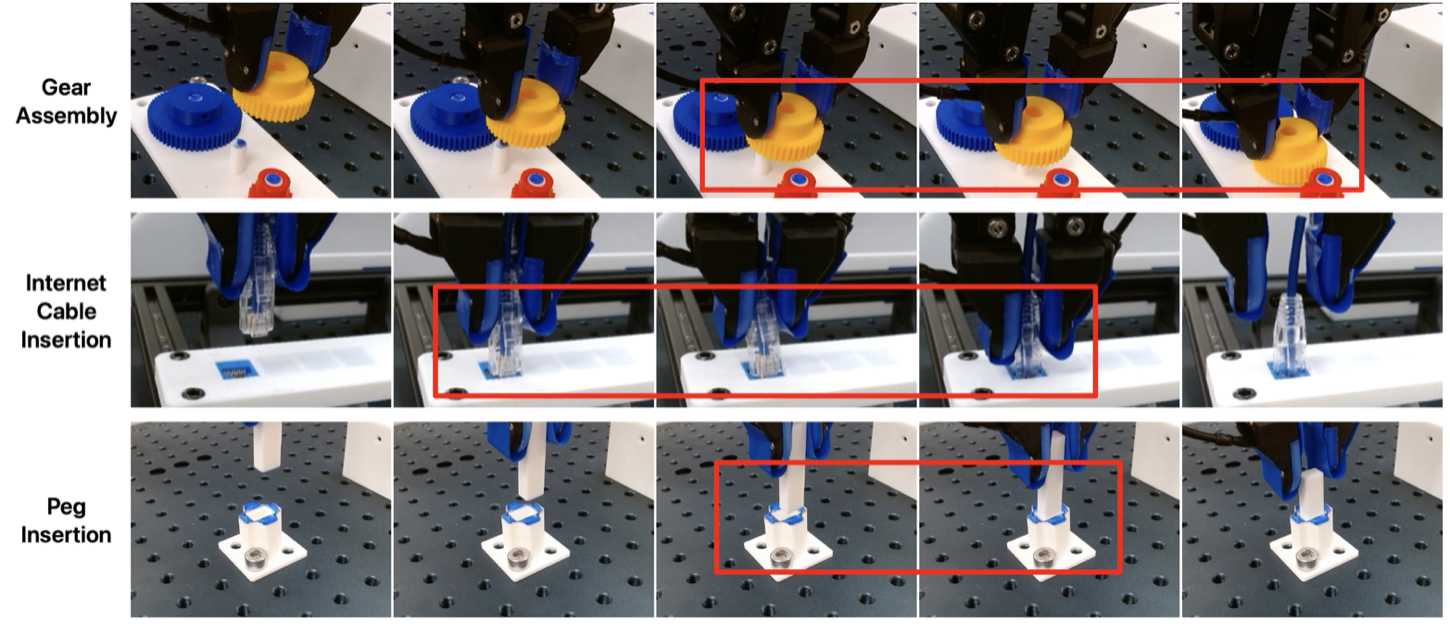}
    \caption{\ABBR{} effectively learns corrective behaviors in real-world experiments. When the initial attempt fails to align the object with the target position, \ABBR{} can still recover under an obstructed visual view by gently maintaining contact with the surrounding surface and using tactile feedback to search for the correct insertion position. This recovery capability substantially improves the success rate over baselines, which often fail to correct the misalignment.
    }
    \label{fig:recovery} 
    \vspace{-0.1in}
\end{figure}

We evaluate \ABBR{} on nine simulation tasks and five real-world tasks against state-of-the-art baselines. Simulation and real-world success rates are reported in \cref{tab:sim_success_rates} and \cref{tab:real_success_rates}, respectively. By leveraging both learned world dynamics and tactile feedback, \ABBR{} exhibits effective corrective behaviors in both simulation and real-world experiments. In the real-world demonstrations, we deliberately include trajectories in which the object is not precisely installed on the first attempt and the robot must perform corrective motions to complete the task. With the limited view of the wrist camera, these corrections are particularly challenging, as the policy must maintain sufficient contact to acquire informative tactile feedback while avoiding damage to the objects or gripper. Benefiting from tactile signal and learned dynamics, \ABBR{} efficiently learns these delicate behaviors and actively attempts recovery when the initial insertion fails, as illustrated in \cref{fig:recovery}.

For example, during peg insertion, the wrist camera view becomes partially occluded by the peg as it approaches the bottom, making it difficult to determine visually whether the peg is aligned with the hole or pressed against its edge. In such cases, \ABBR{} responds to tactile feedback by sliding the peg along the surface toward the hole until successful insertion. Similarly, during gear assembly, the middle gear can become stuck against the other gears and fail to engage properly. \ABBR{} detects this stagnation and performs corrective twisting motions until the gear becomes aligned and fits into place. In contrast, VITA relies solely on visual observations and therefore has limited awareness of contact states. VITA-VT incorporates tactile observations but often fails to learn stable recovery behaviors, tending to become stuck or push blindly after an unsuccessful initial attempt. These results demonstrate that joint action--future modeling encourages the policy representation to capture both visual evolution and local contact dynamics, enabling the fine-grained corrective actions required for contact-rich manipulation.

\subsubsection{Inference Efficiency}

We compare the inference efficiency of \ABBR{} with several baselines. All measurements are conducted on a single NVIDIA RTX 4090 in FP32 with a batch size of 1. Each method generates an action chunk with a horizon of 16, and the reported inference time is averaged over 50 runs. We follow the same sampling settings as in the simulation experiments, using 6 sampling steps for VITA and \ABBR{}, and 100 denoising steps for DP, DP-VT and TAAM. Inference time is measured end-to-end from observation input to action generation, while the corresponding control frequency represents the maximum frequency supported by policy inference. In practical deployment, the actual control frequency may additionally be limited by factors such as the robot control interface and sensor sampling rate. Additionally, for easier comparison, we normalize the metric of VITA to $1\times$ and report the relative values for the other policies.

The results in \cref{tab:model_efficiency} show that \ABBR{} maintains low inference latency comparable to VITA while achieving a control frequency up to 40$\times$ higher than DP. Although \ABBR{} shares the same lightweight flow-matching backbone as VITA, it additionally incorporates tactile encoding and a larger latent space for future prediction. These additions introduce little inference overhead due to the compact model design. Although \ABBR{} operates on a larger latent space, the additional computation is efficiently parallelized on the GPU, resulting in inference latency comparable to VITA. As a result, \ABBR{} remains an agile yet capable World Action Model, making it suitable for high-frequency robot control and potentially for deployment on computationally constrained edge platforms.

\begin{table}[h]
\centering
\caption{Inference efficiency comparison.}
\label{tab:model_efficiency}
\begin{tabular}{lcc|cc}
\toprule
\multirow{2}{*}{\raisebox{-0.8ex}{\textbf{Model}}}
& \multicolumn{2}{c|}{\textbf{Inference Time}}
& \multicolumn{2}{c}{\textbf{Control Frequency}} \\
\cmidrule(lr){2-3}
\cmidrule(lr){4-5}
& \textbf{Value (ms)} & \textbf{Relative}
& \textbf{Value (Hz)} & \textbf{Relative} \\
\midrule

\ABBR{}
& 10.35{\scriptsize$\pm$0.14}
& 1.07$\times$
& \multicolumn{1}{c}{96.62}
& 0.94$\times$ \\
\midrule

VITA
& 9.71{\scriptsize$\pm$0.11}
& 1.00$\times$
& \multicolumn{1}{c}{102.99}
& 1.00$\times$ \\

VITA-VT
& 10.52{\scriptsize$\pm$0.14}
& 1.08$\times$
& \multicolumn{1}{c}{95.06}
& 0.92$\times$ \\

DP
& 408.59{\scriptsize$\pm$0.47}
& 42.08$\times$
& \multicolumn{1}{c}{2.45}
& 0.02$\times$ \\

DP-VT
& 420.34{\scriptsize$\pm$0.36}
& 43.29$\times$
& \multicolumn{1}{c}{2.38}
& 0.02$\times$ \\

TAAM
& 636.34{\scriptsize$\pm$0.90}
& 65.53$\times$
& \multicolumn{1}{c}{1.57}
& 0.02$\times$ \\

\bottomrule
\end{tabular}
\end{table}

\begin{figure}[h]
    \centering
    \includegraphics[width=1.0\linewidth]{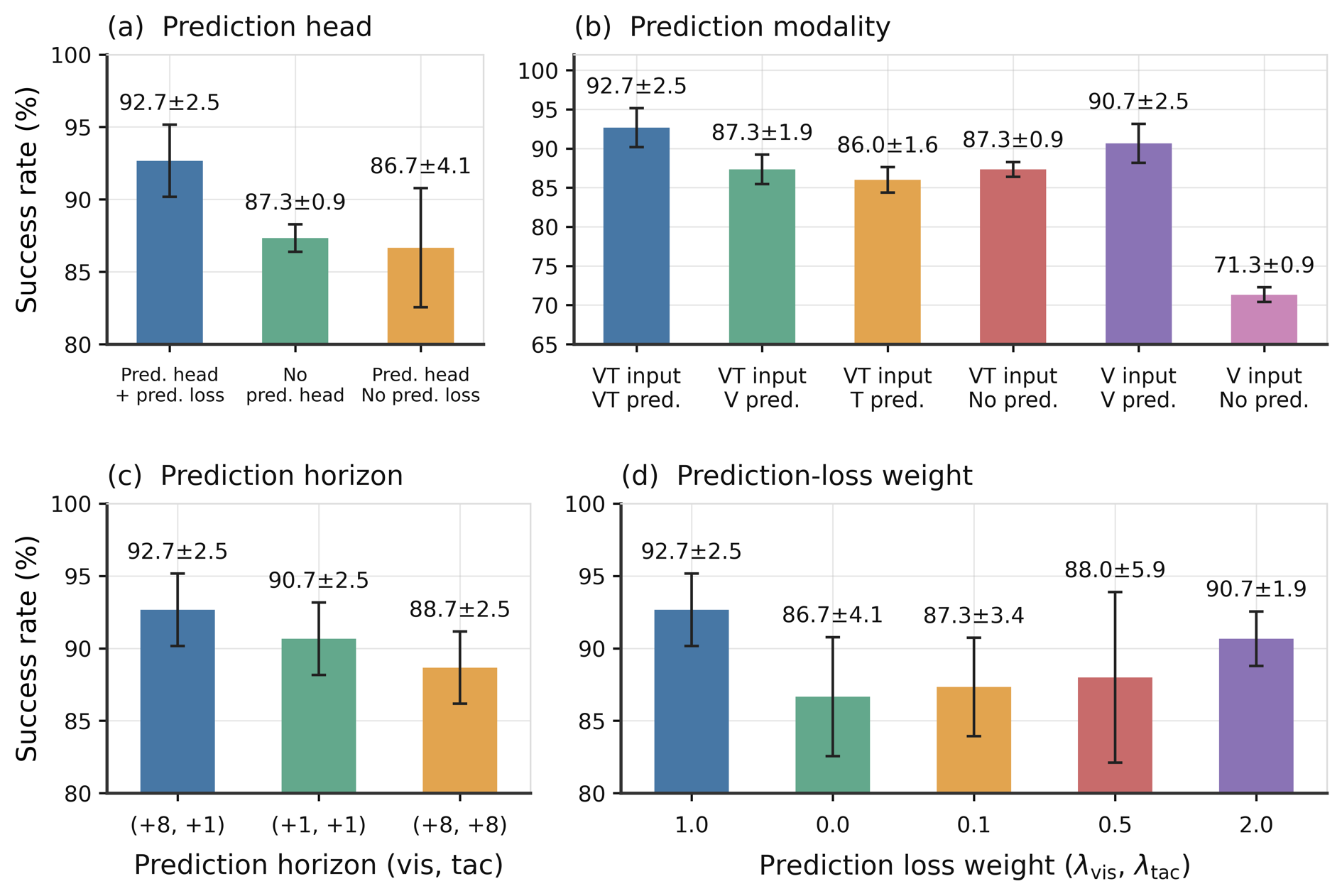}
    \caption{\textbf{Ablation studies of \ABBR{}.}
    (a) Joint action--future modeling improves performance beyond the gain from increased latent capacity.
    (b) Jointly modeling both future visual and tactile observations provides the strongest performance under multimodal input.
    (c) Multi-horizon multimodal prediction better matches the different temporal characteristics of vision and tactile signals.
    (d) We vary the visual and tactile prediction loss weights to explore their respective influence on policy performance.
    }
    \label{fig:ablation} 
    \vspace{-0.1in}
\end{figure}

\subsection{Ablation of World Modeling}
\label{sec: ablation_world_modeling}

We investigate the benefits of world modeling by ablating the prediction losses and the modalities involved in future prediction. As shown in \cref{fig:ablation}(a), \ABBR{} outperforms the baseline without joint action--future modeling. As joint prediction requires a larger flow-matching latent space than the non-predictive baseline, we further disable the world-modeling loss terms $\mathcal{L}_{\mathrm{vis}}$ and $\mathcal{L}_{\mathrm{tac}}$ of \ABBR{} while keeping the enlarged latent for prediction head unchanged. This comparison isolates the effect of world modeling from the additional model capacity. The results show that the improvement mainly comes from joint action--future modeling rather than the larger latent space. By jointly learning actions and future observations, the policy can better capture how its actions affect subsequent states, leading to higher-quality actions that are consistent with the expected future evolution.

We further study the contribution of predicting different observation modalities. In contact-rich manipulation, vision and tactile sensing provide complementary feedback, and modeling their future states can impose different forms of alignment on action learning. We therefore compare different combinations of visual and tactile prediction when both modalities are provided as input, as well as visual prediction when only vision is available. As shown in \cref{fig:ablation}(b), visual world modeling improves performance in the vision-only setting. When both vision and tactile observations are used as input, however, removing the prediction of either modality degrades performance, while \ABBR{} achieves the best result by jointly predicting both future vision and tactile observations. These results suggest that world modeling is most effective when the predicted modalities are consistent with the observation modalities available to the policy. With multimodal input, predicting the future of only one modality provides incomplete supervision of the environment dynamics, whereas jointly modeling both modalities encourages the policy to learn a stronger correspondence among multimodal observations, actions, and future states.

\subsection{Ablation of Prediction Horizon}
We analyze the importance of multi-horizon multimodal prediction by varying the prediction horizons for vision and tactile observations. Visual observations typically evolve smoothly over time, making adjacent-frame prediction highly redundant and potentially too trivial to provide effective supervision to learn visual dynamics. In contrast, tactile signals can change abruptly upon contact, so short-horizon prediction is better suited to capturing fine-grained contact dynamics. We therefore compare our multi-horizon prediction design against two alternatives: next-frame prediction for both vision and tactile, and the same longer prediction horizon for both modalities. As shown in \cref{fig:ablation}(c), using the same prediction horizon for vision and tactile consistently degrades performance. A long prediction horizon for tactile can overlook high-frequency contact changes, whereas next-frame visual prediction provides limited learning signal and fails to capture precise visual dynamics due to the strong redundancy between adjacent frames. Consistent with the observations in \cref{sec: ablation_world_modeling}, ineffective prediction of either modality weakens the alignment between future modeling and multimodal observations, ultimately degrading the quality of the jointly generated actions.

\section{CONCLUSIONS}
\label{sec:conclusion}

This paper presented \ABBR{}, an agile tactile World Action Model that couples multimodal world prediction with action generation for contact-rich manipulation. By using the fused observation latent as the source of a vision-tactile-to-action flow, \ABBR{} jointly generates action, visual, and tactile representations without relying on a large generative backbone or complex conditioning modules. Its multi-horizon multimodal prediction mechanism provides supervision matched to the temporal characteristics of each modality: longer-horizon visual prediction captures meaningful evolution, whereas next-step tactile prediction preserves rapid contact dynamics. Evaluations on nine simulated and five real-world tasks demonstrate strong performance across diverse contact-rich interactions. At the same time, \ABBR{} maintains low inference latency and a high control frequency. Overall, these results demonstrate that tactile World Action Model can be both effective and computationally efficient, making \ABBR{} well suited for precise, high-frequency robot manipulation.







\bibliographystyle{IEEEtran}
\bibliography{references}




\end{document}